\documentclass[10pt,twocolumn,letterpaper]{article}

\usepackage{cvpr}              %

\newcommand{\teaminfo}[3]{
    \begin{table}[H]
        \vspace{-2mm}
        \begin{tabularx}{\linewidth}{@{}lX@{}}
            \toprule
            \textbf{Title:} & {#1} \\
            \textbf{Members:} & {#2} \\
            \textbf{Affiliation:} & {#3} \\
            \bottomrule
        \end{tabularx}
        \vspace{-2mm}
    \end{table}
    }

\usepackage{multirow}
\usepackage{colortbl}
\usepackage{xcolor}
\usepackage{pifont}
\usepackage{makecell}
\usepackage{adjustbox}
\usepackage{threeparttable}
\usepackage{setspace}
\usepackage{tabularx}
\usepackage{float}

\usepackage{pythonhighlight}
\usepackage[linesnumbered,ruled,vlined]{algorithm2e}

\definecolor{cvprblue}{rgb}{0.21,0.49,0.74}
\usepackage[pagebackref,breaklinks,colorlinks,allcolors=cvprblue]{hyperref}

\def\paperID{20} %
\usepackage[misc]{ifsym}
\def\confName{CVPR}
\def\confYear{2026}
\usepackage[accsupp]{axessibility}  %
\usepackage{fontawesome5}
\usepackage{array}
\definecolor{gold}{RGB}{255, 205, 0}
\definecolor{silver}{RGB}{192, 192, 192}
\definecolor{bronze}{RGB}{205, 127, 50}  %
\makeatletter
\newcommand\semilarge{\@setfontsize\semilarge{11.2pt}\@xipt} %
\newcommand\littlelarge{\@setfontsize\littlelarge{11.7pt}\@xipt} %

\makeatother

\title{Report of the 5th PVUW Challenge:\\ Towards More Diverse Modalities in Pixel-Level Understanding}

\author{
    Chang Liu\textsuperscript{*}, \quad 
    Henghui Ding\textsuperscript{*, ${\textrm{\Letter}}$}, \quad 
    Nikhila Ravi\textsuperscript{*}, \quad 
    Yunchao Wei\textsuperscript{*}, \quad 
    Shuting He\textsuperscript{*}, \quad
    Song Bai\textsuperscript{*}, \quad \\
    Philip Torr\textsuperscript{*}, \quad
    Leilei Cao\textsuperscript{*}, \quad 
    \vspace{1.8mm}\\
    Jinrong Zhang, \quad 
    Deshui Miao, \quad
    Xusheng He, \quad 
    Dengxian Gong, \quad
    Zhiyu Wang, \quad
    Mingqi Gao, \quad \\
    Jihwan Hong, \quad
Canyang Wu, \quad 
Weili Guan, \quad 
Jianlong Wu, \quad
Liqiang Nie, \quad
Xingsen Huang, \quad  \\
Yameng Gu, \quad 
Xiaogang Yu, \quad
Xin Li, \quad 
Ming\mbox{-}Hsuan Yang, \quad 
Sijie Li, \quad
Jungong Han, \quad \\
Quanzhu Niu, \quad
Shihao Chen, \quad
Yuanzheng Wu, \quad
Yikang Zhou, \quad
Tao Zhang, \quad
Haobo Yuan, \quad\\
Lu Qi, \quad
Shunping Ji, \quad
Chao Yang, \quad
Chao Tian, \quad
Guoqing Zhu, \quad
Kai Yang, \quad
Zhifan Mo, \quad\\
Haijun Zhang, \quad
Xudong Kang, \quad 
Shutao Li, \quad
Jaeyoung Do
\vspace{1.6mm}\\
\href{https://pvuw.github.io/}{https://pvuw.github.io/}
}

\begin{document}
\maketitle
\renewcommand{\thefootnote}{\fnsymbol{footnote}}
\footnotetext[1]{CVPR 2026 PVUW Challenge organizers. All following authors are
challenge participants from the top-3 teams of each track.}
\footnotetext[0]{${\textrm{\Letter}}$ Corresponding to Henghui Ding (henghui.ding@gmail.com), the Institute of Big Data, Fudan University, Shanghai, China.}

\begin{abstract}
This report summarizes the objectives, datasets, and top-performing methodologies of the 2026 Pixel-level Video Understanding in the Wild (PVUW) Challenge, hosted at CVPR 2026, which evaluates state-of-the-art models under highly unconstrained conditions. To provide a comprehensive assessment, the 2026 edition features three specialized tracks: the MOSE track for tracking objects within densely cluttered and severely occluded scenarios; the MeViS-Text track for localizing targets via motion-focused linguistic expressions; and the newly inaugurated MeViS-Audio track, which pioneers acoustic-driven object segmentation. By introducing previously unreleased challenging data and analyzing the cutting-edge, multimodal solutions submitted by participants, this report highlights the community's latest technical advancements and charts promising future directions for robust video scene comprehension.
\end{abstract}
\vspace{-1em}
    
\section{Introduction}
\label{sec:intro}

Pixel-level video understanding in unconstrained environments continues to be a fundamental challenge in computer vision research~\cite{shuai2024survey,li2024transformer,wu2024towards,ravi2024sam,hesham2025exploiting}. While image-level analysis has achieved remarkable progress~\cite{CCL,BFP,SVC}, understanding dynamic scenes at the pixel level demands more sophisticated approaches that can handle temporal variations, object motion, and complex real-world conditions. Video segmentation tasks, including Video Object Segmentation (VOS) and Referring Video Object Segmentation (RVOS), serve as critical benchmarks for evaluating how well models can track and identify objects across video frames. These capabilities are essential for numerous practical applications, from autonomous vehicles navigating busy streets to robots interacting with dynamic surroundings.

Building upon the success of previous editions~\cite{ding2024pvuw}, we present the 2026 Pixel-level Video Understanding in the Wild (PVUW) workshop and challenge, held in conjunction with CVPR 2026. This year's challenge features three distinct tracks designed to test different aspects of video understanding systems. The first track (MOSE) focuses on complex video object segmentation through the MOSEv2 dataset~\cite{MOSEv2}, where participants tackle scenarios involving occlusions, small objects, adverse weather, and other challenging conditions. The second track (MeViS-Text)~\cite{MeViS} utilizes the MeViS dataset for text-guided video segmentation, requiring models to identify target objects based on linguistic descriptions of their motion and appearance.

A key addition this year is the third track: {audio-based referring video segmentation (MeViS-Audio)}~\cite{MeViS}. This new task challenges participants to segment objects using audio cues rather than text, opening exciting possibilities for multimodal learning and expanding the range of practical applications. By incorporating audio as a guidance signal, this track encourages the development of models that can leverage multiple sensory inputs for more robust video understanding.

The goal of this workshop and challenge series remains to advance the state of the art in video segmentation by providing rigorous benchmarks with real-world data. We continue to expand our datasets with newly collected videos that capture diverse scenarios and challenging conditions. Through these efforts, we aim to boost innovation in algorithms that can reliably perform pixel-level video understanding in the wild, ultimately bringing us closer to perception systems that can operate effectively in complex, unconstrained environments.

\section{The PVUW 2026 Challenge}

This year, besides the two classic tracks MOSE and MeViS~\cite{ding2025pvuw,liu2025lsvos}, we introduce a brand new MeViS-Audio Track, for audio-guided object localization.

\subsection{Three Video Segmentation Tracks}

\indent\textbf{Track 1: MOSE Track.} \textbf{\textit{Complex Video Object Segmentation (MOSEv2)}}~
\cite{MOSE,MOSEv2} targets the problem of tracking and segmenting objects in densely populated and challenging video environments. This track employs the MOSEv2 dataset, which extends the original MOSE benchmark with additional videos captured under difficult conditions including low-light settings, adverse weather, and multi-shot sequences. The dataset comprises 703 video clips with 1,410 annotated objects spanning 28 categories, totaling over 98,000 high-quality masks. These additions further test the robustness of VOS algorithms when faced with camouflaged objects, non-physical entities like shadows, and scenarios requiring external knowledge. The evaluation set incorporates both existing and newly recorded footage, with annotations kept private to ensure fair assessment. This track attracted 76 participating teams, with 38 teams submitting valid results during the testing period. The highest-performing team achieved a $\mathcal{J}\&\mathcal{F}$ score of 88.45\% on the hidden test set.

\vspace{2mm}
\noindent\textbf{Track 2: MeViS-Text Track. } \textit{\textbf{Text-based Motion Expression Video Segmentation (MeViS-Text)}}~
\cite{MeViS,MeViSv2} addresses the task of identifying and segmenting target objects through textual descriptions that emphasize motion characteristics. Built upon the MeViS dataset, this track requires models to interpret language expressions focusing on how objects move rather than merely what they look like. The dataset contains diverse video scenes with multiple interacting objects, where referring expressions explicitly describe motion patterns and temporal dynamics. This formulation poses unique challenges compared to conventional referring segmentation tasks that rely primarily on static visual attributes. The test set for this edition includes fresh video material alongside undisclosed ground-truth annotations. A total of 33 teams enrolled for this track, and 34 teams completed submissions in the final evaluation.

\vspace{2mm}
\noindent\textbf{Track 3 (New): MeViS-Audio Track. }\textit{\textbf{Audio-based Motion Expression Video Segmentation (MeViS-Audio)}}~
\cite{MeViS,MeViSv2} represents a novel direction in multimodal video understanding, where participants must segment objects based on audio descriptions rather than text. This track utilizes synchronized audio narratives that describe object movements and actions, requiring models to process acoustic signals and align them with visual content for accurate segmentation. The introduction of audio guidance opens new avenues for video understanding applications, particularly in scenarios where verbal descriptions may be more natural or accessible than written text. The dataset shares the same video base as MeViS-Text but substitutes textual queries with professionally recorded audio clips. This inaugural edition of the audio track drew 26 registered teams, with 51 submission results. The evaluation follows the same metrics to the MeViS-Text track, enabling direct comparison across different guidance modalities.

\subsection{Evaluation Protocol}

The MOSE track adopts consistent evaluation criteria established in previous PVUW challenges~
\cite{ding2024pvuw,ding2024lsvos}. We measure region similarity ($\mathcal{J}$), contour accuracy ($\mathcal{F}$), and their mean ($\mathcal{J}\&\mathcal{F}$), with the latter serving as the definitive ranking criterion. For two MeViS tracks, with the introduction of No-target samples, we add N-acc. and T-acc. to evaluate no-target performance, following the dataset~\cite{MeViSv2}. The final result is the average of $\mathcal{J}\&\mathcal{F}$, N-acc. and T-acc. Evaluations are performed on the CodaBench platform, which provides automated scoring and leaderboard functionality.

\section{MOSE Track Top Solutions}
\label{sec:mose_method}

\subsection{1st Team in MOSE Track: HITsz\_Dragon}

\teaminfo{Tracking-Enhanced Prompt}{Jinrong Zhang$^{1}$, Canyang Wu$^{1}$, Xusheng He$^{1}$, Weili Guan$^{1,2}$, Jianlong Wu$^{1,2}$, Liqiang Nie$^{1,2}$}{$^1$Harbin Institute of Technology, Shenzhen, China $^2$Shenzhen Loop Area Institute, China}

\begin{figure}[ht]
    \centering
    \includegraphics[width=\linewidth]{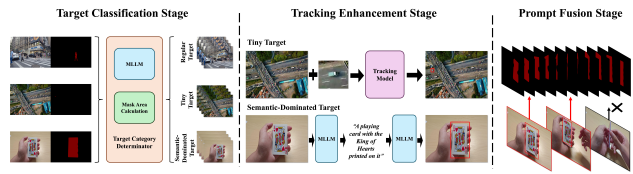}
    \caption{The architecture of the proposed TEP framework, which consists of three main stages: Target Classification, Tracking Enhancement, and Prompt Fusion.}
    \label{fig:TEG}
\end{figure}

The proposed Tracking-Enhanced Prompt (TEP) framework aims to mitigate the inherent limitations of foundation models in highly cluttered environments by dynamically injecting targeted prior knowledge. As shown in Figure \ref{fig:TEG}, the TEP pipeline operates through three sequential stages: Target Classification, Tracking Enhancement, and Prompt Fusion. Initially, a coarse classification mechanism filters video sequences containing special category targets. 
These selected sequences are then routed to the Tracking Enhancement stage, which generates robust, category-specific bounding box prompts. Finally, the Prompt Fusion stage dynamically integrates these auxiliary prompts into the base segmentation model (SAM3) to ensure temporal stability and boost overall segmentation accuracy.

\subsubsection{Target Classification Stage}

In this initial stage, we employ an evaluation metric based on mask area calculations coupled with a Multimodal Large Language Model (MLLM) to categorize targets into three distinct types: {Regular Targets}, {Tiny Targets}, {Semantic-Dominated Targets}.
Videos containing Tiny and Semantic-Dominated targets are forwarded to the subsequent Tracking Enhancement stage.

\subsubsection{Tracking Enhancement Stage}

We deploy two distinct, specialized tracking paradigms to generate reliable bounding box (bbox) coordinates for intermediate frames: {\textbf{Handling Tiny Targets via Tracking Model}}, where we employ SUTrack, an image-prompted tracking methodology. Using images as prompts for detection and tracking can effectively overcome the limitation when targets are difficult to describe with text
\cite{zhang2025just,jiang2024t,oquab2023dinov2}.
{\textbf{Handling Semantic-Dominated Targets via MLLMs}}, where We utilize Qwen3.5. The MLLM first generates a precise textual description of the target based on the reference frame. Subsequently, it performs frame-by-frame object detection guided by this textual prompt, successfully isolating the target from similar instances and outputting the corresponding bounding boxes.

\subsubsection{Prompt Fusion Stage. }
In the final stage, we calculate the Intersection over Union (IoU) between the bounding box of SAM3's predicted mask and the auxiliary bbox provided by our enhancement module. If the IoU falls below a predefined threshold, indicating a potential tracking failure or drift by SAM3, a prompt-switching mechanism is triggered:
\textbf{For Tracking Model Bboxes}: We evaluate the confidence score of the generated bbox. If the confidence is excessively low, the system discards the auxiliary prompt and continues relying on SAM3's inherent pixel diffusion. Conversely, if the confidence score is sufficiently high, the SUTrack bbox is injected as a new, corrective prompt to guide SAM3's subsequent segmentation.
\textbf{For MLLM-Generated Bboxes}: The MLLM takes the visual crops bounded by both SAM3's predicted mask and the MLLM's detection bbox, and compares them against the original target in the first frame. The MLLM evaluates which crop preserves higher semantic and visual fidelity to the reference target. The most accurate bounding box is then selected as the definitive prompt to guide the ongoing VOS process.

\subsection{2nd Team in MOSE Track: tobedone}

\teaminfo{OAMVOS}{Deshui Miao$^{1}$ 
Xingsen Huang$^{1}$ 
Yameng Gu$^{1}$ 
Xiaogang Yu$^{2}$ 
Xin Li$^{1}$ 
Ming\mbox{-}Hsuan Yang$^{3}$}
{$^{1}$Pengcheng Laboratory $^{2}$Guangzhou Hengyan Technology $^{3}$University of California at Merced}

\begin{figure}[ht]
    \centering
    \resizebox{0.9\linewidth}{!}{
        \includegraphics{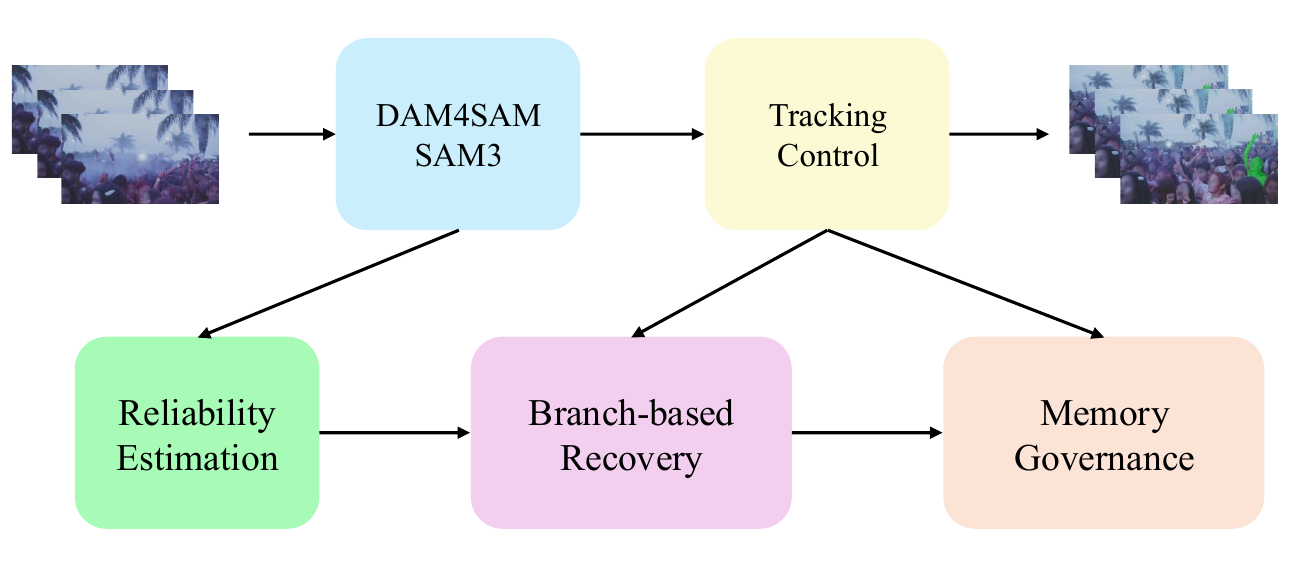}
    }
    \caption{{Pipeline of OAMVOS.}}
    \label{fig:mevis_audio}
     \vspace{-4mm}
\end{figure}

\noindent\textbf{Overview. }
The method is built on top of the SAM3-based DAM4SAM tracker. Let $I_t$ denote frame $t$, $m_t$ the predicted mask, and $p_t \in \mathbb{R}^d$ the corresponding object pointer. After initialization, each frame is processed in one of three modes: {stable}, {ambiguous}, {recovery}. 
In the stable mode, the tracker follows the original DAM4SAM path and performs speculative prediction on the current frame. Reliable predictions are committed as non-conditioning outputs and may later be promoted into DRM. Unreliable predictions trigger an uncertain regime, where several candidate branches are propagated independently, and only a reconfirmed branch is allowed to return to the main path. 

\noindent\textbf{Reliability Estimation in Stable Tracking. }
For every tracked frame, the base predictor returns a primary mask, optional alternative masks, objectness logits, and predicted IoU values. We convert these outputs into four interpretable scores: appearance, motion, geometry, and candidate margin. These scores are designed to capture different aspects of tracking reliability and are combined with the top IoU value to determine whether the current frame should be treated as stable or uncertain.
For details, please refer to our main report.

\noindent\textbf{Branch-Based Recovery. }
Once a frame is marked uncertain, the tracker constructs a branch pool initialized from the current predictor output. A branch may correspond to the primary mask, one of the strongest alternative masks, or an explicit absent-object hypothesis. Each branch owns an independent copy of the inference state, so hypotheses can evolve without polluting the main path. For non-primary masks, the candidate mask is re-injected as a prompt into the predictor to obtain a branch-specific memory state and object pointer. Branch scores accumulate evidence over time rather than relying on a single frame.

\noindent\textbf{Anchor Bank and Delayed DRM Promotion. }
The method maintains two complementary long-term structures: an anchor bank and DRM. The anchor bank stores normalized object pointers from the initialization frame and later high-confidence stable frames. DRM, by contrast, is the stronger conditioning memory already supported by DAM4SAM and is used directly by the transformer encoder during future frame processing.

\noindent\textbf{Conditional Use of Memory Selection. }
Native SAM3 memory selection filters non-conditioning memories using a quality score derived from objectness and predicted IoU. This is useful when recent frames are informative and clean. However, for small-object disappearance and reappearance, recent memories are often dominated by occlusion or near-empty observations, so aggressive filtering can suppress the older anchors needed for recovery.

\noindent\textbf{Conditioning-Memory Attention for Reappearance. }
The base SAM3 tracker limits the number of conditioning frames participating in attention through a parameter $K_c=\texttt{max\_cond\_frames\_in\_attn}$. This is important for efficiency but can hurt long-gap reappearance when the first conditioning frame is discarded in favor of temporally closer DRM entries. We therefore preserve the first conditioning frame explicitly and modestly enlarge the attention budget.

\subsection{3rd Team in MOSE Track: HCVG}

\teaminfo{Re-Prompting SAM 3 via Object Retrieval}{Mingqi Gao\textsuperscript{1}\quad
    Sijie Li\textsuperscript{1}\quad
    Jungong Han\textsuperscript{2}}{ \textsuperscript{1} {School of Computer Science, University of Sheffield} \;
 \textsuperscript{2} {Department of Automation, Tsinghua University}}

\begin{figure}[ht]
    \centering
    \includegraphics[width=\linewidth]{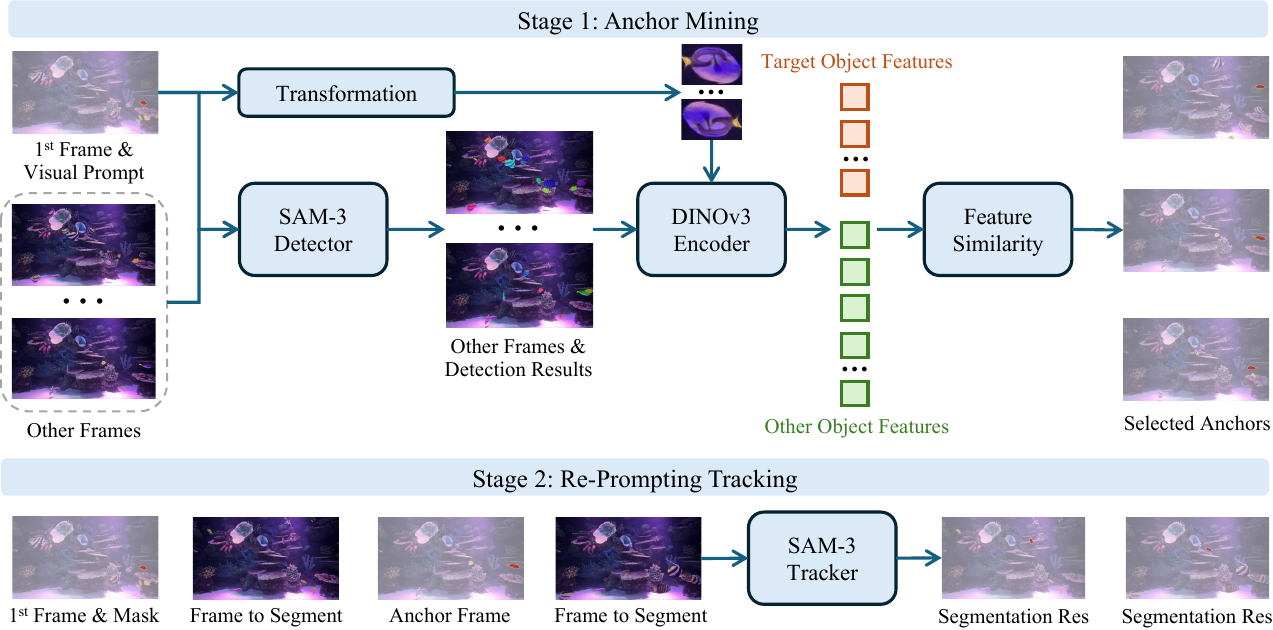}
    \caption{
    Overview of our two-stage framework.
    }
    \label{fig:pipeline}
\end{figure}

We address the MOSEv2 challenge by extending SAM~3 with an automatic re-prompting strategy based on object-level retrieval. The key idea is to move beyond the standard semi-supervised VOS setting where only the first frame serves as the visual anchor. Instead, we automatically identify reliable target candidates from later frames and use them as additional prompts.

\noindent\textbf{Overview. }
As illustrated in Fig.~\ref{fig:pipeline}, our framework consists of two consecutive components: automatic anchor mining and re-prompting-based tracking. 
Our method first searches the video for reliable target instances that can serve as additional anchors, and then uses these anchors together with the first-frame mask to guide subsequent mask propagation. 

\noindent\textbf{Automatic Anchor Mining and Re-Prompting. }
Our method is built on SAM-3~\cite{sam3}. The detector branch can identify and segment objects that belong to the same semantic category as a given visual prompt. 
We first use the provided target mask in the first frame as the initial visual prompt. Based on this prompt, the SAM-3 detector is applied to all remaining video frames to generate candidate objects of the same category as the target, as illustrated in Fig.~\ref{fig:pipeline}. For each candidate, the detector provides both a segmentation mask and the corresponding bounding box.
To distinguish the true target from same-category distractors, we further perform object-level matching using DINOv3 features~\cite{simeoni2025dinov3}. For the target object in the first frame, we crop the object region according to its mask and extract an object-level representation using DINOv3. We also construct transformed versions of the target object, including flipped and rotated variants, and extract DINOv3 features for these transformed views to form a small transformation-aware target feature pool. For each candidate object detected in later frames, we extract its object-level feature in the same way. We then compute the cosine similarity between each candidate and the target feature pool, and use the maximum similarity over the pool as the final matching score. This design improves robustness when the target undergoes large changes in pose, orientation, or appearance.

After matching, we select a few high-confidence candidates as additional anchors. These selected masks are used as reliable pseudo-prompts that complement the original first-frame supervision. In our implementation, we retain top high-similarity candidates while suppressing temporally adjacent redundant selections, so that the final anchors are both reliable and temporally diverse.

Once these anchors are obtained, we inject them back into SAM~3 together with the first-frame ground-truth mask. Concretely, the first frame and the selected anchor frames are provided with masks, while the remaining video frames are fed into the SAM~3 tracker without masks. %

\section{MeViS-Text Track Top Solutions}
\label{sec:mevis_method}

\subsection{1st Team in MeViS-Text Track: HITsz\_Dragon}

\teaminfo{Strong MLLMs Meet SAM3}{Xusheng He$^{1}$, Canyang Wu$^{1}$, Jinrong Zhang$^{1}$, Weili Guan$^{1,2}$, Jianlong Wu$^{1,2}$, Liqiang Nie$^{1,2}$}{$^1$Harbin Institute of Technology, Shenzhen, China $^2$Shenzhen Loop Area Institute, China}

Our method follows a three-stage design. We first transform each target event into a set of instance-level grounding targets through MLLM-based event decomposition. We then apply SAM3-agent on the frame where the target object is most clearly visible to obtain precise seed masks. These seed masks are propagated to the full video by the SAM3 tracker. Finally, we introduce a self-refinement stage that revisits ambiguous predictions and keeps the predicted masks semantically consistent with the original event description.

\noindent \textbf{Stage 1: Event decomposition via key-frame reasoning.}
For each video and target event, we use Gemini-3.1 Pro to analyze the video and decompose the event at the instance level. The model first identifies all valid object instances that truly satisfy the event and isolates the central subject from auxiliary referents. It then selects the frame where each target is most clearly visible and generates a discriminative description for each target. This stage converts a difficult video motion expression into a set of image grounding problems and explicitly separates multiple valid instances for subsequent processing in complex scenes.

\noindent \textbf{Stage 2: SAM3-agent grounding and temporal propagation.} After obtaining the key frame and its discriminative description, we use SAM3-agent to generate a precise seed mask on the selected frame. 
Starting from the selected frame and the target description, the planner analyzes the current result, selects the next SAM3 operation, and updates the decision according to the returned masks until a satisfactory target mask is obtained or the target is judged absent.
During mask selection, the planner can continue to use the description throughout the reasoning process. Once the seed mask is obtained, we initialize the official SAM3 video tracker on the selected frame and propagate it in both temporal directions. For expressions with multiple instances, we segment and track each target independently, and merge the masks only after propagation.

\noindent \textbf{Stage 3: Self-refinement by consistency checking.}

We begin by inspecting the predicted masks, where empty predictions and highly overlapping masks for distinct targets indicate that the original description is not sufficiently discriminative. We then use Qwen3.5-Plus\cite{qwen35blog} to regenerate a more precise description and rerun the same SAM3-agent grounding procedure. In addition, for expressions with directional or negative constraints, we further perform a behavior-level verification step by sampling frames from the video, highlighting the predicted mask boundaries, and asking an MLLM to judge whether the tracked object is truly consistent with the original event description. If a prediction fails this consistency check, we send it back to the refinement branch for another round of re-grounding. This refinement stage improves the robustness of the overall pipeline.

\subsection{2nd Team in MeViS-Text Track: Still Awesome SaSaSa2VA}

\teaminfo{SaSaSaSa2VA}{ Dengxian Gong$^{1}$,
    Quanzhu Niu$^{1}$,
    Shihao Chen$^{1}$,
    Yuanzheng Wu$^{1}$,
    Yikang Zhou$^{1}$,
    Tao Zhang$^{1}$,
    Haobo Yuan$^{2}$,
    Lu Qi$^{1}$,
    Shunping Ji$^{1}$}{$^{1}$Wuhan University 
    $^{2}$University of California, Merced}

\begin{figure}[ht]
  \centering
  \includegraphics[width=0.97\linewidth]{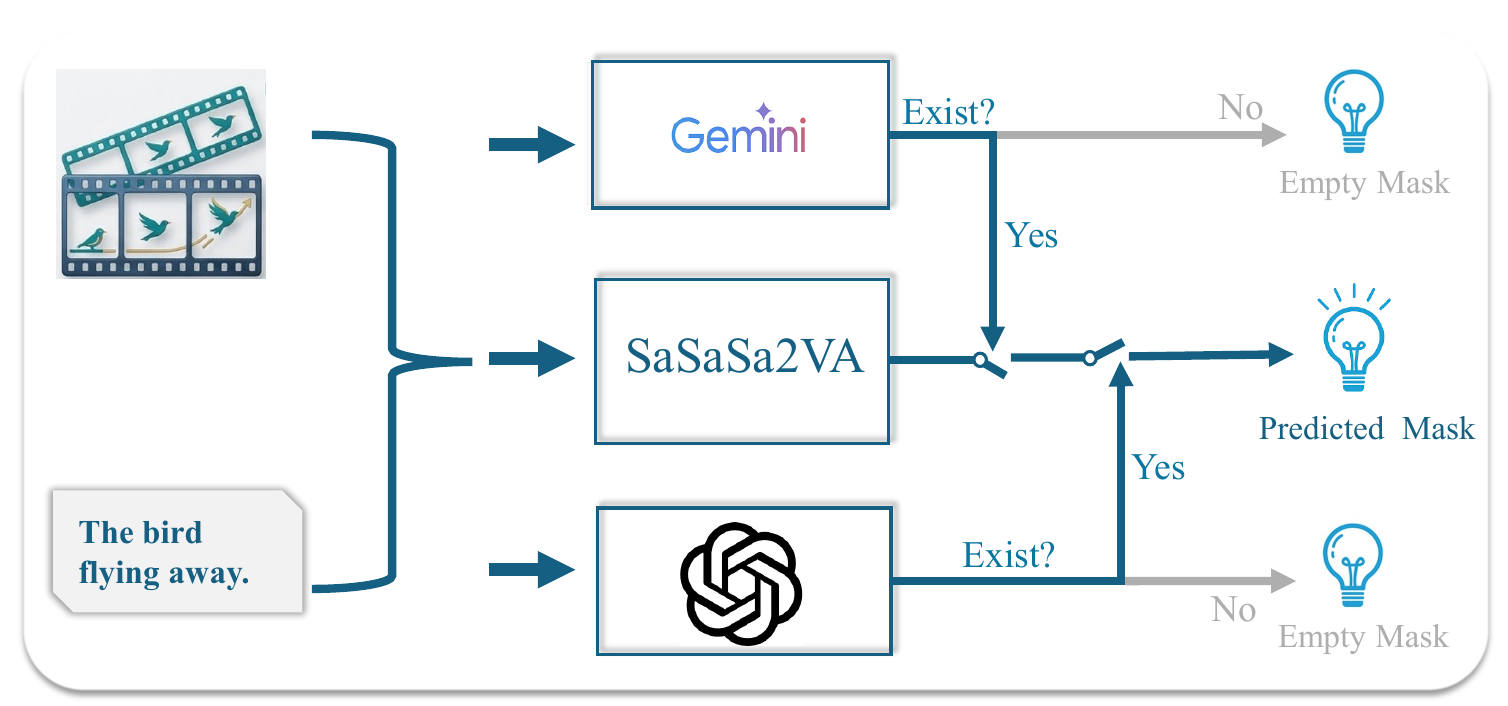}
  \caption{{The Existence-aware verification illustration of our method.} }
  \label{fig:ea}
\end{figure}

\label{sec:method_sa2va}

\noindent\textbf{Overview.} As illustrated in~\cref{fig:ea}, we adopt a target existence-aware verification strategy conditioned on video-language inputs to pre-determine the existence of the referred target in the video and then refine the model predictions accordingly.

\noindent\textbf{Existence-aware verification.}
To rectify the inherent positive bias within the SaSaSa2VA base model, we leverage the high-level semantic reasoning capabilities of state-of-the-art closed-source models, namely \emph{Gemini 3-Flash-Preview}~\cite{gemini_pro} and \emph{GPT-5.4}~\cite{gpt54} as the pre-inference safeguard. Specifically, for each video-expression pair $(\mathcal{V}, \mathcal{T})$, these models function as a dual-consensus jury: all frames $\mathbf{I}_t$ along with the referring expression $\mathcal{T}$ are fed into both models to evaluate the target's presence. We categorize an expression as `null-target' only under a unanimous consensus, where both models independently confirm the object's absence. In such instances, our framework bypasses the standard segmentation pipeline and preemptively returns a null masklet $\mathbf{M} = \mathbf{0}$. This strategic `consensus gating' effectively shields the system from producing forced-mapping hallucinations, thereby substantially rehabilitating the $\mathrm{N\text{-}acc}$ metric and improving the overall reliability of the segmentation output.

\noindent\textbf{Test-time Augmentation}
While SaSaSa2VA~\cite{niu20251st} employs a heavy ensemble mechanism across two dimensions-averaging predictions from multiple sampling strategies (e.g., uniform sampling, content-aware, cyclic) and models of varying scales-our solution adopts a significantly more simple inference pipeline. Specifically, we dispense with both the multi-strategy voting and multi-model aggregation, opting exclusively for a single-model approach powered by the \emph{Uniform+} sampling strategy. For video sequences with an original duration shorter than the training constraint $T$, we maintain temporal coverage by assigning dual \texttt{[SEG]} tokens to specific frames near the clip boundaries. The final segmentation is then derived by averaging the masks from these two corresponding tokens. By focusing on this singular, high-efficiency configuration, we substantially reduce the computational cost while maintaining robust mask generation capabilities.

\subsection{3rd Team in MeViS-Text Track: tobedone}

\teaminfo{AgentRVOS}{Deshui Miao$^{1}$,
Chao Yang$^{1}$,
Chao Tian$^{1}$,
Guoqing Zhu$^{1}$,
Kai Yang$^{2}$,
Zhifan Mo$^{3}$,
Xin Li$^{1}$
}
{$^{1}$Pengcheng Laboratory 
$^{2}$Wuhan Textile University 
$^{3}$Yixiang Innovation Technology (Shenzhen)}

Our method follows a coarse-to-fine design that decouples semantic grounding, temporal propagation, and decision-level refinement. The key idea is to retain Sa2VA as the primary semantic interpreter of the language query, while introducing lightweight agentic modules to explicitly reason about target presence, anchor reliability, and frame-level ambiguity. The overall pipeline consists of five stages: presence verification, coarse Sa2VA segmentation, anchor extraction from Sa2VA predictions, SAM3-based propagation, and planner-guided conflict resolution.

\textbf{Stage 1: Presence Agent. }
The first stage is a \emph{Presence Agent}, whose sole purpose is to determine whether the target referred to by $q$ actually exists in the video. Formally, the agent predicts the binary decision$
e = \Psi_{\mathrm{pres}}(V,q)$,
where $\Psi_{\mathrm{pres}}$ denotes the presence judgment module.

If $e=0$, the system terminates early and outputs zero masks for all frames. This design prevents the downstream segmentation model from hallucinating object masks for non-existent targets, which is a common failure mode in open-ended language-conditioned video understanding. By separating existence reasoning from dense segmentation, the proposed framework reduces false positives and avoids unnecessary refinement on invalid queries.

\textbf{Stage 2: Coarse Segmentation. }
When the Presence Agent predicts $e=1$, we invoke Sa2VA on the full video and the referring expression to obtain an initial coarse segmentation trajectory:
$
\tilde{\mathcal{M}}
=\{\tilde{m}_t\}_{t=1}^{T}
=\mathrm{Sa2VA}(V,q).$
Here, $\tilde{m}_t$ denotes the coarse mask predicted for frame $I_t$. In our implementation, Sa2VA directly processes the entire video and returns frame-aligned predictions through \texttt{predict\_forward}, yielding a full-video segmentation hypothesis without requiring additional frame sampling or external temporal stitching.

\textbf{Stage 3: Anchor Selection. }
The third stage extracts reliable anchor frames from the Sa2VA trajectory. The purpose of this step is to transform Sa2VA's coarse semantic predictions into structured prompts that can initialize a stronger propagation model.

Let $\mathcal{A}\subseteq\{1,\dots,T\}$ denote the selected anchor set. For each anchor frame $a\in\mathcal{A}$, we derive geometric prompts from the corresponding Sa2VA mask $\tilde{m}_a$. In particular, we compute a bounding box
$b_a=\mathrm{BBox}(\tilde{m}_a),$
and, if needed, additional point prompts extracted from the mask support region. These prompts are then passed to SAM3 as initialization signals.

\textbf{Stage 4: SAM3 Propagation. }
Starting from the Sa2VA-derived anchors, we employ SAM3 to propagate the object masks over time and obtain a refined trajectory
$\mathcal{T}=\{(\hat{m}_t,\hat{b}_t)\}_{t=1}^{T},$
where $\hat{m}_t$ and $\hat{b}_t$ denote the propagated mask and bounding box at frame $t$, respectively.

\textbf{Stage 5: Planner-Based Selection on Ambiguous Frames. }
We introduce a lightweight \emph{Planner} that operates as a local decision module rather than a global controller. Specifically, for ambiguous frames, the Planner compares two candidate mask sources: the original Sa2VA prediction $\tilde{m}_t$ and the propagated SAM3 prediction $\hat{m}_t$.
Based on visual similarity, local temporal context, and prediction reliability, the Planner selects the more trustworthy candidate for each ambiguous frame. Therefore, its role is not to conduct exhaustive search over the video, but to resolve local conflicts between semantic grounding and geometric propagation.

\section{MeViS-Audio Track Top Solutions}
\label{sec:mevis_audio_method}

\subsection{1st Team in MeViS-Audio Track: tobedone}

\teaminfo{APRVOS}{Deshui Miao$^{1}$,
Yameng Gu$^{1}$,
Chao Yang$^{1}$,
Xin Li$^{1}$,
Haijun Zhang$^{2}$,
Ming\mbox{-}Hsuan Yang$^{3}$}{$^{1}$Pengcheng Laboratory 
$^{2}$Harbin Institute of Technology 
$^{3}$University of California at Merced }

\noindent\textbf{Stage 1: VibeVoice-ASR~\cite{peng2026vibevoice} for Audio-to-Text Conversion}
VibeVoice-ASR~\cite{peng2026vibevoice} is suitable for this setting because it is designed for long-form speech recognition and can provide structured transcriptions that preserve useful contextual cues, such as speaker turns, temporal alignment, and utterance content.

\noindent\textbf{Stage 2: Visual Existence Judgment}
After obtaining the ASR transcript, the next stage performs visual existence judgment. 
We employ Qwen3-VL~\cite{Qwen2.5-VL} as a visual judge. Given the transcript-derived referring phrase and a set of sampled video frames, the module estimates whether the described entity can be visually grounded in the scene. The result is stored as \texttt{presence\_info.target\_exists}.

\noindent\textbf{Stage 3: Prompt Construction for Sa2VA~\cite{yuan2025sa2va}}
If the target is judged to be present, the transcript is converted into a segmentation-oriented textual prompt that matches the input format expected by Sa2VA~\cite{yuan2025sa2va}. In the current dataset loader, this is achieved through one of two prompt templates:\texttt{<image>\textbackslash nPlease segment \{exp\}.}; \texttt{<image>\textbackslash n\{exp\} Please respond with a segmentation mask.}
The second template is used when the expression already resembles a question.

\noindent\textbf{Stage 4: Coarse Semantic Segmenter}
The constructed prompt, together with the full video, is then fed into Sa2VA. Let
$\tilde{\mathcal{M}} = \{\tilde{m}_t\}_{t=1}^{T} = \mathrm{Sa2VA}(V, q_{\text{asr}})$
denote the coarse mask trajectory returned by \texttt{predict\_forward}. Since Sa2VA processes the entire video and outputs \texttt{prediction\_masks} aligned with the original frame sequence, it functions as a full-video segmenter rather than a frame-local grounding model.

\noindent\textbf{Stage 5: Agentic Verification}
This stage is where the complexity of audio-conditioned Ref-VOS is handled through explicit reasoning rather than being hidden inside one end-to-end segmentation score. The agent can analyze the transcript quality, infer the most relevant temporal window, identify candidate anchor frames, and decompose the query into positive constraints, negative constraints, and temporal hints. A planner may decide which refinement strategy is most appropriate; scout modules may search for frames in which Sa2VA provides the most trustworthy localization signal; and a critic may assess whether the current trajectory is semantically plausible and temporally coherent.

\noindent\textbf{Stage 6: Refinement from Trusted Anchors}
Once the agent identifies a reliable anchor frame, the corresponding Sa2VA mask can be converted into geometric prompts for SAM3-based refinement. For a trusted frame $a$, we derive
$b_a = \mathrm{BBox}(\tilde{m}_a),
p_a = \mathrm{Center}(\tilde{m}_a),$
or, if needed, an alternative refinement point predicted by a visual-language model. These geometric prompts are then used to initialize SAM3, which propagates the target both forward and backward in time.

\subsection{2nd Team in MeViS-Audio Track: HNU-VPAI}

\teaminfo
    {ASR-SaSaSa2VA}
    {Zhiyu Wang, 
    Xudong Kang, 
    Shutao Li}
    {Hunan University}

\begin{figure}[ht]
  \centering
  \includegraphics[width=1.0\linewidth]{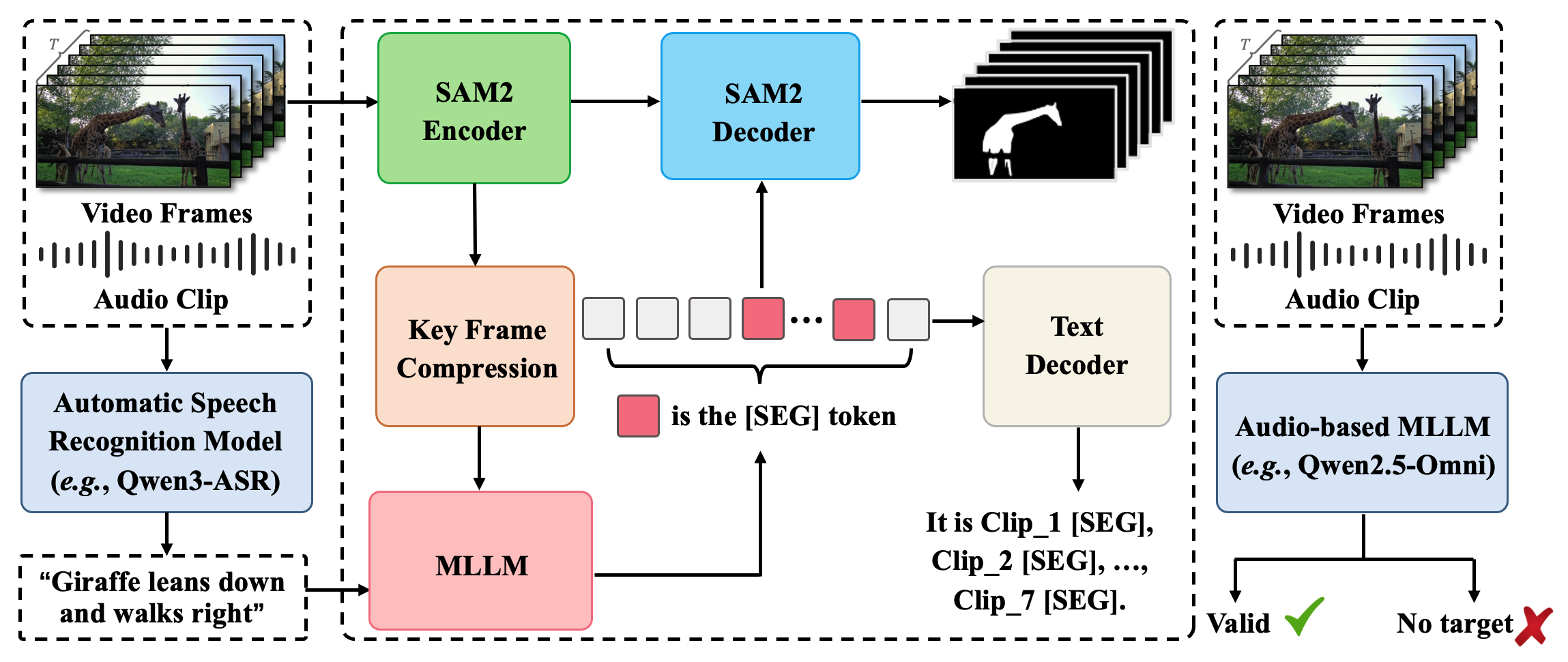}
  \caption{{Overview of our method.}}
  \label{fig:sa}
\end{figure}

In this section, we present \textbf{ASR-SaSaSa2VA}, a modular framework for audio-guided video object segmentation. As illustrated in Fig.~\ref{fig:sa}, our method decomposes the task into three stages: (1) converting audio inputs into textual motion descriptions via automatic speech recognition (ASR) models, (2) performing text-based referring video segmentation using a pre-trained model, and (3) filtering out invalid queries through a no-target expression detection module.

\noindent\textbf{Automatic Speech Recognition. }
Given an input video $\mathcal{V}$ and its associated audio signal $\mathcal{A}$, our first step is to convert the audio modality into a textual representation. Specifically, we employ an off-the-shelf automatic speech recognition (ASR) model to transcribe $\mathcal{A}$ into a text sequence $\mathcal{T}$:
In this work, we adopt \textbf{Qwen3-ASR}~\cite{shi2026qwen3} as our ASR backbone.

\noindent\textbf{Text-based Video Segmentation. }
Given the video $\mathcal{V}$ and the transcribed text $\mathcal{T}$, we adopt a pre-trained text-based referring video segmentation model to predict pixel-level masks of the target object. In this work, we build upon {SaSaSa2VA}~\cite{niu20251st}, which extends the Sa2VA framework~\cite{yuan2025sa2va} with enhanced temporal modeling capability.
SaSaSa2VA improves upon Sa2VA by addressing its limitations in temporal modeling. In particular, Sa2VA processes only a small number of sampled frames and relies on a single \texttt{[SEG]} token to represent the entire video, which restricts its ability to capture long-range temporal dependencies.

\noindent\textbf{No-target Expression Detection. }
In this challenge, not all audio inputs correspond to valid target objects in the video. To improve robustness, we introduce a no-target expression detection module that determines whether a given audio clip contains a valid referring expression.

Specifically, we adopt \textbf{Qwen2.5-Omni} as the backbone, an end-to-end multimodal large language model capable of jointly modeling text, images, audio, and video. Benefiting from its unified multimodal representation and strong cross-modal reasoning ability, Qwen2.5-Omni is naturally well-suited for understanding the semantic alignment between audio descriptions and visual content. 
To adapt the model to our task, we fine-tune Qwen2.5-Omni using a parameter-efficient LoRA strategy to perform binary classification: $y = f_{\text{cls}}(\mathcal{A}, \mathcal{V})$,
where $y \in \{0,1\}$ indicates whether the audio describes a target object present in the video.

If $y = 0$, the input is classified as a no-target expression, and the segmentation stage is skipped. Otherwise, the transcribed text $\mathcal{T}$ is passed to the segmentation model. This lightweight adaptation enables effective filtering of ambiguous or irrelevant audio inputs, reducing false positives and improving the overall robustness of the system without introducing significant computational overhead.

\subsection{3rd Team in MeViS-Audio Track: SNU-AIDAS}

\teaminfo
    {VIRST-Audio}
    {Jihwan Hong$^{1}$, Jaeyoung Do$^{1,2}$}
    {AIDAS Laboratory, $^1$IPAI \& $^2$ECE, Seoul National University}

\begin{figure}[ht]
  \centering
    \includegraphics[width=\linewidth]{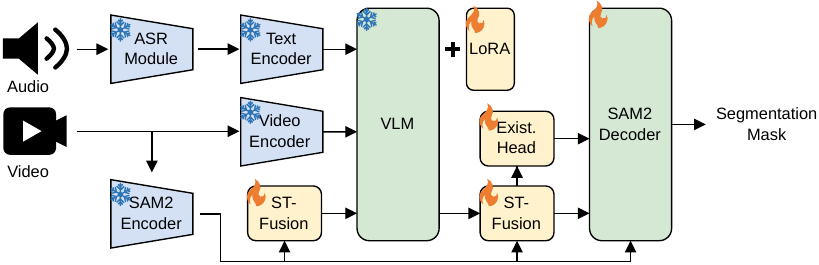} 
    \caption{{Overall architecture of VIRST-Audio.}}
    \label{fig:overall_arch}
\end{figure}

We propose \textbf{VIRST-Audio}, which builds upon VIRST (Video-Instructed Reasoning Assistant for Spatio-Temporal Segmentation)~\cite{hong2026virst}. The overall pipeline is illustrated in Fig.~\ref{fig:overall_arch}. 
VIRST is a VLM-based segmentation framework that combines global semantic understanding with pixel-level segmentation. It integrates CLIP~\cite{cherti2023openclip}-based video encoder features and SAM2~\cite{ravi2024sam} encoder features via an ST-Fusion module~\cite{hong2026virst}, which performs cross-attention using a learnable token \texttt{[ST]}. The fused representation is then used as a prompt for the SAM2 mask decoder.
To extend this framework to ARVOS, VIRST-Audio incorporates an ASR (Automatic Speech Recognition) module that converts the input audio into text. The transcribed text is then used as the language input to the VLM. 

\noindent\textbf{Existence-Aware Segmentation Gating. }
Robust segmentation in the absence of the target object is a critical challenge in RVOS, as false positives can significantly limit real-world applicability and incur unnecessary computational cost. Recent works~\cite{yan2024visa,li2023r2vos,MeViSv2} have increasingly focused on mitigating false positive and false negative predictions. To address this issue, we introduce an \textbf{existence-aware gating mechanism} that determines whether the audio-referred target object exists in the video and suppresses spurious segmentation when the target is absent.
For this, we define an indicator function that determines whether the referred target object is present in the video. From the ST-Fusion model output $\mathbf{F} \in \mathbb{R}^{N \times T \times D}$, which is used as prompts for the mask decoder, we apply a lightweight existence prediction module:
$z = f_{\text{exist}}(\mathbf{F}), p = \sigma(z),$
where $p$ denotes the probability that the referred target exists in the video.

For training, we supervise the existence prediction using a binary cross-entropy (BCE) loss, where the target label is defined based on whether the ground-truth mask is non-empty.
At inference time, the predicted existence probability $p$ is compared with a threshold $\tau$ to determine whether segmentation should be performed. If $p < \tau$, the model directly outputs an empty prediction without invoking the segmentation module. Otherwise, segmentation is conducted using the predicted prompts and propagated across the video.

\section{Conclusion and Discussion}
\label{sec:conclusion}

The 2026 iteration of the PVUW workshop witnessed remarkable enthusiasm and widespread participation across its three competitive tracks. This robust involvement underscores the community's escalating demand for dependable perception systems capable of operating in unconstrained, real-world environments. By analyzing the leading solutions submitted this year, we can extract several prominent technological trends.

First, the synergy between meticulously annotated benchmarks and vision foundation models is undeniable. Comprehensive datasets such as MOSEv2 and the MeViS suite provide the necessary complexity to test modern algorithms thoroughly. In response, a majority of the top-tier teams built their pipelines upon advanced architectures, notably SAM 2 and SAM 3, adapting them for intricate temporal propagation and robust target tracking. Second, the integration of Multimodal Large Language Models (MLLMs) has shifted from a supplementary feature to a core necessity, especially within the text and newly introduced audio tracks. We observed a significant paradigm shift where models like Qwen and Gemini were deployed not merely for semantic interpretation, but as crucial agents for "existence verification" and spatial-temporal reasoning, effectively curbing hallucinated masks and false positive predictions.

Moving forward, our organizational team is dedicated to continuously refreshing and diversifying the evaluation sets for both the MOSE and MeViS benchmarks. By introducing novel challenges and expanding multi-sensory modalities, we strive to consistently drive the frontier of dynamic, pixel-level scene comprehension in the wild.

\clearpage
\footnotesize{\paragraph{Acknowledgement.} This work was supported by the National Natural Science Foundation of China (NSFC) under Grant No. 62472104 and the Science and Technology Commission of Shanghai Municipality (No.~25511103600).

{
    \small
    \bibliographystyle{ieeenat_fullname}
    \bibliography{main}

\begin{thebibliography}{33}
\providecommand{\natexlab}[1]{#1}
\providecommand{\url}[1]{\texttt{#1}}
\expandafter\ifx\csname urlstyle\endcsname\relax
  \providecommand{\doi}[1]{doi: #1}\else
  \providecommand{\doi}{doi: \begingroup \urlstyle{rm}\Url}\fi

\bibitem[Bai et~al.(2025)Bai, Chen, Liu, Wang, Ge, Song, Dang, Wang, Wang, Tang, Zhong, Zhu, Yang, Li, Wan, Wang, Ding, Fu, Xu, Ye, Zhang, Xie, Cheng, Zhang, Yang, Xu, and Lin]{Qwen2.5-VL}
Shuai Bai, Keqin Chen, Xuejing Liu, Jialin Wang, Wenbin Ge, Sibo Song, Kai Dang, Peng Wang, Shijie Wang, Jun Tang, Humen Zhong, Yuanzhi Zhu, Mingkun Yang, Zhaohai Li, Jianqiang Wan, Pengfei Wang, Wei Ding, Zheren Fu, Yiheng Xu, Jiabo Ye, Xi Zhang, Tianbao Xie, Zesen Cheng, Hang Zhang, Zhibo Yang, Haiyang Xu, and Junyang Lin.
\newblock Qwen2.5-vl technical report.
\newblock \emph{arXiv preprint arXiv:2502.13923}, 2025.

\bibitem[Carion et~al.(2026)Carion, Gustafson, Hu, Debnath, Hu, Suris, Ryali, Alwala, Khedr, Huang, Lei, Ma, Guo, Kalla, Marks, Greer, Wang, Sun, Rädle, Afouras, Mavroudi, Xu, Wu, Zhou, Momeni, Hazra, Ding, Vaze, Porcher, Li, Li, Kamath, Cheng, Dollár, Ravi, Saenko, Zhang, and Feichtenhofer]{sam3}
Nicolas Carion, Laura Gustafson, Yuan-Ting Hu, Shoubhik Debnath, Ronghang Hu, Didac Suris, Chaitanya Ryali, Kalyan~Vasudev Alwala, Haitham Khedr, Andrew Huang, Jie Lei, Tengyu Ma, Baishan Guo, Arpit Kalla, Markus Marks, Joseph Greer, Meng Wang, Peize Sun, Roman Rädle, Triantafyllos Afouras, Effrosyni Mavroudi, Katherine Xu, Tsung-Han Wu, Yu Zhou, Liliane Momeni, Rishi Hazra, Shuangrui Ding, Sagar Vaze, Francois Porcher, Feng Li, Siyuan Li, Aishwarya Kamath, Ho~Kei Cheng, Piotr Dollár, Nikhila Ravi, Kate Saenko, Pengchuan Zhang, and Christoph Feichtenhofer.
\newblock Sam 3: Segment anything with concepts.
\newblock In \emph{ICLR}, 2026.

\bibitem[Cherti et~al.(2023)Cherti, Beaumont, Wightman, Wortsman, Ilharco, Gordon, Schuhmann, Schmidt, and Jitsev]{cherti2023openclip}
Mehdi Cherti, Romain Beaumont, Ross Wightman, Mitchell Wortsman, Gabriel Ilharco, Cade Gordon, Christoph Schuhmann, Ludwig Schmidt, and Jenia Jitsev.
\newblock Reproducible scaling laws for contrastive language-image learning.
\newblock In \emph{CVPR}, pages 2818--2829, 2023.

\bibitem[Ding et~al.(2018)Ding, Jiang, Shuai, Liu, and Wang]{CCL}
Henghui Ding, Xudong Jiang, Bing Shuai, Ai~Qun Liu, and Gang Wang.
\newblock Context contrasted feature and gated multi-scale aggregation for scene segmentation.
\newblock In \emph{Proceedings of the IEEE conference on computer vision and pattern recognition}, pages 2393--2402, 2018.

\bibitem[Ding et~al.(2019{\natexlab{a}})Ding, Jiang, Liu, Thalmann, and Wang]{BFP}
Henghui Ding, Xudong Jiang, Ai~Qun Liu, Nadia~Magnenat Thalmann, and Gang Wang.
\newblock Boundary-aware feature propagation for scene segmentation.
\newblock In \emph{Proceedings of the IEEE/CVF international conference on computer vision}, pages 6819--6829, 2019{\natexlab{a}}.

\bibitem[Ding et~al.(2019{\natexlab{b}})Ding, Jiang, Shuai, Liu, and Wang]{SVC}
Henghui Ding, Xudong Jiang, Bing Shuai, Ai~Qun Liu, and Gang Wang.
\newblock Semantic correlation promoted shape-variant context for segmentation.
\newblock In \emph{Proceedings of the IEEE/CVF Conference on Computer Vision and Pattern Recognition}, pages 8885--8894, 2019{\natexlab{b}}.

\bibitem[Ding et~al.(2023{\natexlab{a}})Ding, Liu, He, Jiang, and Loy]{MeViS}
Henghui Ding, Chang Liu, Shuting He, Xudong Jiang, and Chen~Change Loy.
\newblock {MeViS}: A large-scale benchmark for video segmentation with motion expressions.
\newblock In \emph{Proceedings of the IEEE/CVF International Conference on Computer Vision}, pages 2694--2703, 2023{\natexlab{a}}.

\bibitem[Ding et~al.(2023{\natexlab{b}})Ding, Liu, He, Jiang, Torr, and Bai]{MOSE}
Henghui Ding, Chang Liu, Shuting He, Xudong Jiang, Philip~HS Torr, and Song Bai.
\newblock {MOSE}: A new dataset for video object segmentation in complex scenes.
\newblock In \emph{Proceedings of the IEEE/CVF International Conference on Computer Vision}, pages 20224--20234, 2023{\natexlab{b}}.

\bibitem[Ding et~al.(2024{\natexlab{a}})Ding, Hong, Liu, Xu, Yang, Fan, Miao, Gu, Li, He, et~al.]{ding2024lsvos}
Henghui Ding, Lingyi Hong, Chang Liu, Ning Xu, Linjie Yang, Yuchen Fan, Deshui Miao, Yameng Gu, Xin Li, Zhenyu He, et~al.
\newblock {LSVOS} challenge report: Large-scale complex and long video object segmentation.
\newblock In \emph{ECCV Workshop}, 2024{\natexlab{a}}.

\bibitem[Ding et~al.(2024{\natexlab{b}})Ding, Liu, Wei, Ravi, He, Bai, Torr, Miao, Li, He, et~al.]{ding2024pvuw}
Henghui Ding, Chang Liu, Yunchao Wei, Nikhila Ravi, Shuting He, Song Bai, Philip Torr, Deshui Miao, Xin Li, Zhenyu He, et~al.
\newblock {PVUW} 2024 challenge on complex video understanding: Methods and results.
\newblock In \emph{ECCV Workshop}, 2024{\natexlab{b}}.

\bibitem[Ding et~al.(2025{\natexlab{a}})Ding, Liu, He, Ying, Jiang, Loy, and Jiang]{MeViSv2}
Henghui Ding, Chang Liu, Shuting He, Kaining Ying, Xudong Jiang, Chen~Change Loy, and Yu-Gang Jiang.
\newblock Mevis: A multi-modal dataset for referring motion expression video segmentation.
\newblock \emph{IEEE Transactions on Pattern Analysis and Machine Intelligence}, 2025{\natexlab{a}}.

\bibitem[Ding et~al.(2025{\natexlab{b}})Ding, Liu, Ravi, He, Wei, Bai, and Torr]{ding2025pvuw}
Henghui Ding, Chang Liu, Nikhila Ravi, Shuting He, Yunchao Wei, Song Bai, and Philip Torr.
\newblock Pvuw 2025 challenge report: Advances in pixel-level understanding of complex videos in the wild.
\newblock In \emph{Proceedings of the Computer Vision and Pattern Recognition Conference}, pages 2669--2678, 2025{\natexlab{b}}.

\bibitem[Ding et~al.(2025{\natexlab{c}})Ding, Ying, Liu, He, Jiang, Jiang, Torr, and Bai]{MOSEv2}
Henghui Ding, Kaining Ying, Chang Liu, Shuting He, Xudong Jiang, Yu-Gang Jiang, Philip~HS Torr, and Song Bai.
\newblock {MOSEv2}: A more challenging dataset for video object segmentation in complex scenes.
\newblock \emph{arXiv preprint arXiv:2508.05630}, 2025{\natexlab{c}}.

\bibitem[{Google DeepMind}(2026)]{gemini_pro}
{Google DeepMind}.
\newblock Gemini 3.1 pro: Best for complex tasks and bringing creative concepts to life, 2026.

\bibitem[Hesham et~al.(2025)Hesham, Liu, Sun, Ding, Yang, Konukoglu, Geng, and Jiang]{hesham2025exploiting}
Syed Ariff~Syed Hesham, Yun Liu, Guolei Sun, Henghui Ding, Jing Yang, Ender Konukoglu, Xue Geng, and Xudong Jiang.
\newblock Exploiting temporal state space sharing for video semantic segmentation.
\newblock In \emph{Proceedings of the IEEE/CVF Conference on Computer Vision and Pattern Recognition}, 2025.

\bibitem[Hong and Do(2026)]{hong2026virst}
Jihwan Hong and Jaeyoung Do.
\newblock Virst: Video-instructed reasoning assistant for spatiotemporal segmentation.
\newblock In \emph{CVPR}, 2026.
\newblock to appear.

\bibitem[Jiang et~al.(2024)Jiang, Li, Zeng, Ren, Liu, and Zhang]{jiang2024t}
Qing Jiang, Feng Li, Zhaoyang Zeng, Tianhe Ren, Shilong Liu, and Lei Zhang.
\newblock T-rex2: Towards generic object detection via text-visual prompt synergy.
\newblock In \emph{ECCV}, pages 38--57. Springer, 2024.

\bibitem[Li et~al.(2023)Li, Wang, Xu, Li, Raj, and Lu]{li2023r2vos}
Xiang Li, Jinglu Wang, Xiaohao Xu, Xiao Li, Bhiksha Raj, and Yan Lu.
\newblock Towards robust referring video object segmentation with cyclic structural consensus.
\newblock In \emph{ICCV}, 2023.

\bibitem[Li et~al.(2024)Li, Ding, Yuan, Zhang, Pang, Cheng, Chen, Liu, and Loy]{li2024transformer}
Xiangtai Li, Henghui Ding, Haobo Yuan, Wenwei Zhang, Jiangmiao Pang, Guangliang Cheng, Kai Chen, Ziwei Liu, and Chen~Change Loy.
\newblock Transformer-based visual segmentation: A survey.
\newblock \emph{IEEE transactions on pattern analysis and machine intelligence}, 2024.

\bibitem[Liu et~al.(2025)Liu, Ding, Ying, Hong, Xu, Yang, Fan, Gao, Chen, Miao, et~al.]{liu2025lsvos}
Chang Liu, Henghui Ding, Kaining Ying, Lingyi Hong, Ning Xu, Linjie Yang, Yuchen Fan, Mingqi Gao, Jingkun Chen, Yunqi Miao, et~al.
\newblock Lsvos 2025 challenge report: Recent advances in complex video object segmentation.
\newblock \emph{arXiv preprint arXiv:2510.11063}, 2025.

\bibitem[Niu et~al.(2025)Niu, Gong, Chen, Zhang, Zhou, Yuan, Qi, Li, and Ji]{niu20251st}
Quanzhu Niu, Dengxian Gong, Shihao Chen, Tao Zhang, Yikang Zhou, Haobo Yuan, Lu Qi, Xiangtai Li, and Shunping Ji.
\newblock The 1st solution for 7th lsvos rvos track: Sasasa2va.
\newblock \emph{arXiv preprint arXiv:2509.16972}, 2025.

\bibitem[{OpenAI}(2026)]{gpt54}
{OpenAI}.
\newblock Introducing gpt-5.4, 2026.

\bibitem[Oquab et~al.(2023)Oquab, Darcet, Moutakanni, Vo, Szafraniec, Khalidov, Fernandez, Haziza, Massa, El-Nouby, et~al.]{oquab2023dinov2}
Maxime Oquab, Timoth{\'e}e Darcet, Th{\'e}o Moutakanni, Huy Vo, Marc Szafraniec, Vasil Khalidov, Pierre Fernandez, Daniel Haziza, Francisco Massa, Alaaeldin El-Nouby, et~al.
\newblock Dinov2: Learning robust visual features without supervision.
\newblock \emph{arXiv preprint arXiv:2304.07193}, 2023.

\bibitem[Peng et~al.(2026)Peng, Yu, Chang, Wang, Dong, Hao, Tu, Yang, Wang, Xu, et~al.]{peng2026vibevoice}
Zhiliang Peng, Jianwei Yu, Yaoyao Chang, Zilong Wang, Li Dong, Yingbo Hao, Yujie Tu, Chenyu Yang, Wenhui Wang, Songchen Xu, et~al.
\newblock Vibevoice-asr technical report.
\newblock \emph{arXiv preprint arXiv:2601.18184}, 2026.

\bibitem[Ravi et~al.(2024)Ravi, Gabeur, Hu, Hu, Ryali, Ma, Khedr, R{\"a}dle, Rolland, Gustafson, et~al.]{ravi2024sam}
Nikhila Ravi, Valentin Gabeur, Yuan-Ting Hu, Ronghang Hu, Chaitanya Ryali, Tengyu Ma, Haitham Khedr, Roman R{\"a}dle, Chloe Rolland, Laura Gustafson, et~al.
\newblock {SAM 2}: Segment anything in images and videos.
\newblock \emph{arXiv preprint arXiv:2408.00714}, 2024.

\bibitem[Shi et~al.(2026)Shi, Wang, Guo, Wang, Zhang, Zhang, Guo, Hao, Xi, Yang, et~al.]{shi2026qwen3}
Xian Shi, Xiong Wang, Zhifang Guo, Yongqi Wang, Pei Zhang, Xinyu Zhang, Zishan Guo, Hongkun Hao, Yu Xi, Baosong Yang, et~al.
\newblock Qwen3-asr technical report.
\newblock \emph{arXiv preprint arXiv:2601.21337}, 2026.

\bibitem[Shuai et~al.(2024)Shuai, Ding, Ma, Tu, Jiang, and Tao]{shuai2024survey}
Xincheng Shuai, Henghui Ding, Xingjun Ma, Rongcheng Tu, Yu-Gang Jiang, and Dacheng Tao.
\newblock A survey of multimodal-guided image editing with text-to-image diffusion models.
\newblock \emph{arXiv:2406.14555}, 2024.

\bibitem[Sim{\'e}oni et~al.(2025)Sim{\'e}oni, Vo, Seitzer, Baldassarre, Oquab, Jose, Khalidov, Szafraniec, Yi, Ramamonjisoa, et~al.]{simeoni2025dinov3}
Oriane Sim{\'e}oni, Huy~V Vo, Maximilian Seitzer, Federico Baldassarre, Maxime Oquab, Cijo Jose, Vasil Khalidov, Marc Szafraniec, Seungeun Yi, Micha{\"e}l Ramamonjisoa, et~al.
\newblock Dinov3.
\newblock \emph{arXiv preprint arXiv:2508.10104}, 2025.

\bibitem[Team(2026)]{qwen35blog}
Qwen Team.
\newblock Qwen3.5: Accelerating productivity with native multimodal agents, 2026.

\bibitem[Wu et~al.(2024)Wu, Li, Xu, Yuan, Ding, Yang, Li, Zhang, Tong, Jiang, et~al.]{wu2024towards}
Jianzong Wu, Xiangtai Li, Shilin Xu, Haobo Yuan, Henghui Ding, Yibo Yang, Xia Li, Jiangning Zhang, Yunhai Tong, Xudong Jiang, et~al.
\newblock Towards open vocabulary learning: A survey.
\newblock \emph{IEEE Transactions on Pattern Analysis and Machine Intelligence}, 46\penalty0 (7):\penalty0 5092--5113, 2024.

\bibitem[Yan et~al.(2024)Yan, Wang, Yan, Jiang, Hu, Kang, Xie, and Gavves]{yan2024visa}
Cilin Yan, Haochen Wang, Shilin Yan, Xiaolong Jiang, Yao Hu, Guoliang Kang, Weidi Xie, and Efstratios Gavves.
\newblock Visa: Reasoning video object segmentation via large language models.
\newblock In \emph{European Conference on Computer Vision}, pages 98--115. Springer, 2024.

\bibitem[Yuan et~al.(2025)Yuan, Li, Zhang, Huang, Xu, Ji, Tong, Qi, Feng, and Yang]{yuan2025sa2va}
Haobo Yuan, Xiangtai Li, Tao Zhang, Zilong Huang, Shilin Xu, Shunping Ji, Yunhai Tong, Lu Qi, Jiashi Feng, and Ming-Hsuan Yang.
\newblock Sa2va: Marrying sam2 with llava for dense grounded understanding of images and videos.
\newblock \emph{arXiv preprint arXiv:2501.04001}, 2025.

\bibitem[Zhang et~al.(2025)Zhang, Wang, Liu, Liu, Jin, Zhang, Meng, and Hu]{zhang2025just}
Jinrong Zhang, Penghui Wang, Chunxiao Liu, Wei Liu, Dian Jin, Qiong Zhang, Erli Meng, and Zhengnan Hu.
\newblock Just a few glances: Open-set visual perception with image prompt paradigm.
\newblock In \emph{AAAI}, pages 9969--9976, 2025.

\end{thebibliography}
}

\end{document}